\documentclass[twocolumn,conference]{IEEEtran}
\usepackage[T1]{fontenc}
\usepackage[latin9]{inputenc}
\usepackage{color}
\usepackage{amsmath}
\usepackage{amssymb}
\usepackage{graphicx}
\usepackage[unicode=true,
 bookmarks=true,bookmarksnumbered=true,bookmarksopen=true,bookmarksopenlevel=1,
 breaklinks=false,pdfborder={0 0 0},pdfborderstyle={},backref=false,colorlinks=true]
 {hyperref}
\hypersetup{pdftitle={Estimation of higher order SDEs},
 pdfauthor={Stijn de Waele},
 pdfpagelayout=OneColumn, pdfnewwindow=true, pdfstartview=XYZ, plainpages=false}

\makeatletter
\usepackage[caption=false,font=footnotesize]{subfig}
\usepackage{pgfplots}

\makeatother

\begin{document}
\title{Accurate Characterization of Non-Uniformly Sampled Time Series using
Stochastic Differential Equations}
\author{\IEEEauthorblockN{Stijn de Waele}\IEEEauthorblockA{}ExxonMobil Research
and Engineering Company, Annandale, New Jersey, USA.}
\maketitle
\begin{abstract}
Non-uniform sampling arises when an experimenter does not have full
control over the sampling characteristics of the process under investigation.
Moreover, it is introduced intentionally in algorithms such as Bayesian
optimization and compressive sensing. We argue that Stochastic Differential
Equations (SDEs) are especially well-suited for characterizing second
order moments of such time series. We introduce new initial estimates
for the numerical optimization of the likelihood, based on incremental
estimation and initialization from autoregressive models. Furthermore,
we introduce model truncation as a purely data-driven method to reduce
the order of the estimated model based on the SDE likelihood. We show
the increased accuracy achieved with the new estimator in simulation
experiments, covering all challenging circumstances that may be encountered
in characterizing a non-uniformly sampled time series. Finally, we
apply the new estimator to experimental rainfall variability data.
\end{abstract}

\section{Introduction}

Non-uniformly sampled time series are found in a wide range of applications.
They typically occur when the experimenter has limited control over
signal sampling. Sampling times may be determined by a natural process,
for example in turbulent flow characterization using laser-Doppler
anemometry \cite{damaschke2018ldaspectra}, analysis of climate time
series \cite{thirumalai2020climate} and astronomy \cite{khan2017exoplanets}.
Alternatively, non-uniform sampling can occur because of an irregular
human sampling, e.g. in power systems sensor data \cite{stankovic2017power},
oil production surveillance \cite{tewari201production} and vital
signs measurements in medical applications \cite{tan2020medical,barbieri2020icureadmission}.
Finally, non-uniform sampling is a key attribute of certain algorithms,
including Bayesian optimization \cite{ghahramani2015bayesianopt}
and compressive sensing \cite{turunctur2019compressive}.

It has been shown theoretically that alias-free spectral estimates
can be obtained using non-uniform sampling\cite{shapiro1960aliasfree}.
As most theoretical results, this result is asymptotic, i.e., it holds
in the limit where the number of observations $N$ tends to infinity.
The result can easily be understood intuitively, as follows: As a
signal is sampled at random times for a long enough time, any desired
number of observations is a available at an arbitrarily low sampling
interval. Hence, the spectrum can be estimated alias-free up to an
arbitrary high frequency. Following the same intuition, it is clear
that this asymptotic result breaks down for finite $N$, because the
shortest time interval will be finite in that case.

The objective of this work is to accurately characterize the second
order moments of a non-uniformly sampled stationary stochastic process,
which can be expressed in terms of the power spectral density in the
frequency domain. Given the close correspondence between the error
in the log power spectrum and the Kullback-Leibler discrepancy \cite{broersen2006automatic},
we will use ``spectral estimation'' as a shorthand for the aforementioned
objective. In the time domain, the second order moments are expressed
using the covariance function, or, equivalently, the kernel of a Gaussian
processes \cite{rasmussen2003gaussian}. 

Existing approaches include non-parametric methods, such as the Lomb-Scargle
spectral estimate \cite{springford2020lombscargle}, and the slotting
technique for estimation of the covariance function \cite{rehfeld2011comparison}.
Parametric techniques include discrete-time autoregressive models
\cite{broersen2007beyondrate} and stochastic differential equations
with random initial estimates \cite{kelly2014astronomy}. While parametric
techniques show the most promising results, previous work also report
inaccurate spectral spectral estimates at higher frequencies for higher-order
models, and sensitivity to initial conditions of the maximum likelihood
(ML) fitting procedure \cite{jones1984sensitive-init,broersen2006sensitive-init},
which limit the practical use of these algorithms.

Our main contribution is to propose a new algorithm for spectral estimation
using Stochastic Differential Equations (SDEs) based on incremental
parameter estimation and data-driven model truncation, that has been
validated in simulation experiments. We explain how model truncation
in SDEs results greater accuracy than discrete-time models through
analyzing the SDE(1) case. For the problem of spectral estimation
from non-uniformly sampled time series, asymptotic theory provides
a poor description of actual behavior. Hence, simulation experiments
are indispensable to establish the accuracy of an estimator, and are
therefore a key component of this paper.

In section \ref{sec:SDE-motivation}, we motivate the usage of SDEs
for spectral estimation and provide a number of basic definitions
and results. Section \ref{sec:MLE} describes the new SDE parameter
estimation procedure. Section \ref{sec:simulations} contains the
design of the simulation experiments, which is used in section \ref{sec:truncation}
to quantify the error reduction achieved through model truncation,
and in section \ref{sec:performance} for the overall evaluation of
the proposed estimator. Finally, we apply the estimator to experimental
rainfall variability data in section \ref{sec:rainfall}.

\section{\label{sec:SDE-motivation}Stochastic Differential Equations: Motivation
and definitions}

We use stochastic differential equations (SDEs) to estimate the power
spectral density of a time series. We motivate this choice over available
alternatives as follows:
\begin{itemize}
\item By using \emph{a parametric model }we can formulate of a maximum likelihood
(ML) estimator. Unlike many non-parametric estimators, ML estimators
have desirable theoretical properties such as asymptotic efficiency.
Parametric spectral estimators have been most successful at achieving
the benefit of high-frequency spectra in finite samples \cite{broersen2007beyondrate},
whereas non-parametric methods such as the Lomb-Scargle estimator
\cite{springford2020lombscargle} have very high variance, which mostly
limits their use to finding a single peak in the spectrum under high
signal-to-noise conditions. Similarly, in the time domain, correlation
estimated with the non-parametric slotting technique violate the positive-definiteness
property that is required for a valid autocorrelation function \cite{rehfeld2011comparison}.
Parametric models can achieve the same flexibility as non-parametric
estimates by increasing the model order.
\item The SDE model is a \emph{continuous time} model, which can operate
directly on non-uniform samples without the need to introduce a regular
grid as is required for discrete-time estimates. Moreover, estimated
discrete time time series models may have poles that do not have a
continuous-time counterpart \cite{soderstrom1991contarma}.
\end{itemize}
The stochastic differential equation of order $p$ for the process
$y$ as a function of time $t$ is given by:
\begin{equation}
\frac{d^{p}y}{dt^{p}}+\sum_{i=1}^{p}a_{i}\frac{d^{i-1}y}{dt}=\epsilon\label{eq:sde}
\end{equation}
where $a_{i}$ are the SDE coefficients, collectively denoted $\mathbf{a}$,
and $\epsilon$ is continuous Gaussian white noise with standard deviation
$\sigma_{\epsilon}$.

The SDE model can be rewritten as an equivalent state-space model:
\begin{align}
\text{\ensuremath{\frac{d\mathbb{\mathbf{z}}}{dt}}} & =\mathbf{A}\mathbf{z}+\mathbf{\epsilon}\label{eq:state-space}\\
y & =C\text{\ensuremath{\mathbf{z}}}\nonumber 
\end{align}
where the state $\mathbf{z}$ and $\mathbb{\epsilon}$ are $p$-dimensional
time series and $C\mathbf{z}=z_{1}$. The matrix $\mathbf{A}$ and
the covariance matrix $\mathbf{Q}$ of $\mathbb{\epsilon}$ can be
computed from $\left\{ \mathbf{a},\sigma_{\epsilon}\right\} $, see
\cite{jones1981fitcar}.

The SDE process has a power spectral density given by \cite{sarkka2019sde}:
\[
h_{y}\left(f\right)=\mathbf{C}\left(\mathbf{A}-i2\pi f\mathbf{I}\right)^{-1}\mathbf{Q}\left(\mathbf{A}+i2\pi f\mathbf{I}\right)^{-T}\mathbf{C}^{T}
\]
and the covariance function at lag $\tau>0$ of:
\begin{equation}
R\left(\tau\right)=\mathbf{\mathbf{P}}_{s}\exp\left(\mathbf{A}t\right)^{T}\label{eq:autocov}
\end{equation}
where $\mathbf{\mathbf{P}}_{s}$ is the stationary covariance matrix.

The state space equation may be diagonalized to represent the SDE
as:
\begin{align*}
\text{\ensuremath{\frac{d\mathbb{\mathbf{z}}'}{dt}}} & =\Lambda\mathbb{\mathbf{z}}'+\mathbf{\epsilon'}\\
y= & C'\text{\ensuremath{\mathbf{z}}}'
\end{align*}
where the eigenvalues are equal to the the roots $r_{i}$ (collectively
denoted $\mathbf{r}$) of the characteristic equation of (\ref{eq:sde})
for $\epsilon=0$, and $C'\text{\ensuremath{\mathbf{z}}}'=\sum z'_{i}$.
We refer to this parameterization as the \textbf{roots parameterization},
while we refer to the $\left\{ \mathbf{a},\sigma_{\epsilon}\right\} $
parameterization as the \textbf{coefficients} \textbf{parameterization}. 

The key advantage of the roots parameterization is that stationarity
can be expressed as the requirement that the real part of the roots
is negative : $\textrm{Re}\left(r_{i}\right)<0$. Furthermore, the
computational complexity of the likelihood computation is more efficient
for large model order $p$.

Despite the advantages, many researchers use the coefficient representation.
This may be motivated by the reduced computational complexity for
lower-order models. In addition, SDEs can be used to parameterize
a kernel as part of a larger machine learning model, e.g. in deep
learning models \cite{chen2018neuralode} or for posterior sampling
using a probabilistic program \cite{carpenter2017stan}. Currently
many major automatic differentiation packages such as PyTorch \cite{paszke2017pytorch-autodiff}
and the Stan autodiff library \cite{carpenter2015stan-autodiff} do
not support complex-valued parameters, therefore necessitating the
use of the coefficient representation. Therefore, we consider both
representations in this work.

\section{\label{sec:MLE}Maximum likelihood estimation}

\subsection{Likelihood computation}

The exact log likelihood $L$ is computed recursively using the Kalman
filtering equations. Process stationarity is exploited for the first
observation, i.e., it has the stationary covariance matrix $\mathbf{P}_{s}$.
Given a value for the SDE parameters $\mathbf{a}$ or $\mathbf{r}$,
the analytical expression for the Maximum Likelihood standard deviation
$\sigma_{\epsilon}$ is used \cite{jones1981fitcar}. For unstable
models, or errors due to the finite machine precision, a log likelihood
of $L=-\infty$ is produced.

The computation largely follows cited algorithms, with the following
improvements to increase computational efficiency:
\begin{enumerate}
\item For the coefficient representation, we follow \cite{sarkka2019sde}
for the measurement and likelihood\textbf{ }steps. The prediction
step computes the conditional mean $\mathbf{\mu}_{n|n-1}$ and covariance
matrix $\mathbf{P}_{n|n-1}$ of the state $\mathbf{z}_{n}$:
\begin{eqnarray*}
\mathbf{\mu}_{n|n-1} & = & \mathbf{F}\mathbf{\mu}_{n-1}\\
\mathbf{P}_{n|n-1} & = & \mathbf{F}\mathbf{P}_{n-1}\mathbf{F}^{T}+\int_{\tau=0}^{\Delta t}\exp(\mathbf{A}\tau)\mathbf{Q}\exp(\mathbf{A}\tau)^{T}d\tau
\end{eqnarray*}
\end{enumerate}
where $\mathbf{F}=\exp\left(\mathbf{A}t\right)$. To compute the integral,
instead of using the Matrix Fraction Decomposition proposed in \cite{sarkka2019sde},
we eliminate the covariance matrix $\mathbf{Q}$ to yield:
\begin{align*}
\mathbf{P}_{n|n-1} & =\mathbf{P}_{s}-\mathbf{F}(\mathbf{P}_{s}-\mathbf{P}_{n-1})\mathbf{F}^{T}
\end{align*}

\begin{enumerate}
\item [2] For the root representation, we use the algorithm in \cite{jones1981fitcar},
with the following alternative computation for the stationary covariance
matrix $\mathbf{P}'_{s}$:
\[
\left[\mathbf{P}'_{s}\right]_{ij}=-\frac{\left[\mathbf{Q}'_{s}\right]_{ij}}{\left(r_{i}+\bar{r_{j}}\right)},
\]
which is a corollary of eq. 30 in \cite{jones1981fitcar} for $t_{k}-t_{k-1}\rightarrow\infty$.
\end{enumerate}
Given the extensive literature on Kalman filtering, these improvements
may have been previously reported in the literature. We still report
them here to accurately represent to algorithms used in this work.

\subsection{Optimization}

Optimization of the likelihood is performed using the Limited-Memory
BFGS algorithm \cite{nocedal2006optimization}, with derivatives obtained
through automatic differentiation. For the coefficient representation,
a necessary condition for stability is that all coefficients $a_{i}$
are positive; it is also sufficient for $p\le2$ \cite{mattuck2011diffeq-stability}.
To improve optimization results, parameter values are constrained
to a configurable interval: $a_{i}\in\left\langle a_{l},a_{h}\right\rangle $.
Wide limits should be set to allow for a wide range of models. As
an indication, $a_{l}$ is related to the duration $D$ of the time
series $a_{l}\lesssim1/D$, and $a_{h}$ is related to the shortest
sampling interval $\Delta t_{m}$, $a_{h}\gtrsim1/\Delta t_{m}$.
In the presented simulation results, we use $a_{i}\in\left\langle 10^{-3},10^{3}\right\rangle $.
In the roots representation, the real and imaginary part of the roots
are similarly constrained.

\subsection{Initialization\label{subsec:Initialization}}

Accurate initialization of the optimization is key to achieving high-quality
estimates. The sensitivity of SDE parameter estimation to initial
conditions has been acknowledged in the literature \cite{jones1984sensitive-init},
in particular for higher-order models \cite{broersen2006sensitive-init}.

The first element in the algorithm is \textbf{incremental estimation}:
The estimate for the SDE($p$) model is initiated from a lower order
SDE($p'$) model ($p'<p$). The motivation for incremental estimation
is the observation that per parameter, lower order models often have
the largest contribution to the model fit. The lower order models
are expanded by adding a single random real root (for $p'=p-1$) or
a conjugate pair of random complex roots (for $p'=p-2$) to the lower
order model. An alternative for a this root initialization would be
to use a large initial value, as this corresponds to a model that
is most similar to the lower-order model. However, this extreme initialization
does not result in successful convergence to a finite value during
the numerical optimization.

The second element is \textbf{initiation from autoregressive }(AR)
\textbf{models }estimated from resampled data using the Burg estimator
\cite{percival2020spectral}. Resampling is performed using nearest
neighbor interpolation and linear interpolation. Intentionally, basic
interpolation methods are used here, because more advanced interpolation
methods tend to produce artificially smooth signals, or, in the frequency
domain, a power spectrum with a very large dynamic range. This results
in less accurate models, since the estimators attempts to fit the
artificially introduced low power spectral density at high frequencies,
at the expense of modeling the actual process dynamics \cite{broersen2006automatic}.
AR roots that do not have a corresponding continuous-time root \cite{soderstrom1991contarma}
are replaced by randomly generated roots.

The usage of autoregressive models is similar to the methodology proposed
in \cite{broersen2009spurious} for reducing the variance in spectral
estimates based on AR models. However, our work is distinct in the
following aspects: (i) we do not need to introduce an arbitrary criterion
to remove roots in the upper half of the spectrum, which could eliminate
true spectral peaks, and (ii) we only use the AR model as an initial
estimate, allowing further optimization of the likelihood during numerical
optimization.

These main components are supplemented with purely random initialization
\cite{kelly2014astronomy,goodfellow2016deep} and a truncation phase
that occurs after incremental estimation, where successively lower
model orders are initiated from the most significant roots of higher
order models.

\subsection{Implementation}

The estimator is implemented in Julia 1.4, using \href{https://julianlsolvers.github.io/Optim.jl/stable/}{Optim.jl}
\cite{mogensen2018optim} for L-BFGS optimization. Gradients are computed
with automatic differentiation using \href{https://fluxml.ai/Zygote.jl/latest/}{Zygote.jl}
\cite{innes2019zygote}. Julia was used because it combines an expressive
syntax with high execution speed. Zygote is a package for automatic
differentiation that supports all functions used in the likelihood
computation, notably including the matrix exponential \cite{branvcik2008expadjoint},
and supports complex-valued parameters.

\section{\label{sec:simulations}Design of experiments}

\subsection{\label{subsec:Test-processes}Test processes}

The experiment is designed to cover the following process characteristics
that are challenging for parameter estimation from non-uniformly sampled
data:
\begin{enumerate}
\item \textbf{Overfit}: When the order of the estimated model matches the
order that of the generating process, accurate models can be estimated
\cite{broersen2007beyondrate}. However, the additional flexibility
of a higher order parameters can lead to large errors, more so than
the small statistical error that is observed in parameter estimation
from regularly sampled data.
\item \textbf{High dynamic range}: Estimation of the spectral density at
frequencies where the true density is low is challenging for many
estimators, due to a phenomenon similar to spectral leakage \cite{bos2002autoregressive}.
\item \textbf{Spectral details beyond average sampling rate}: While asymptotic
theory predicts alias-free estimates, capturing spectral details at
higher frequency remains challenging in practice, because limited
information high-frequency information is available in finite samples.
\item \textbf{Model misspecification}: Any estimation procedure should continue
to work well when the actual process cannot be described exactly using
the estimated model structure.
\end{enumerate}
To cover these characteristics, we use the following test processes:
\begin{itemize}
\item \textbf{Case A}: SDE(1) process with parameter $a_{1}=-1/200$. Covariance
function: $R\left(\tau\right)=\exp\left(a_{1}\tau\right)$. Addresses
characteristics \#1 and \#2.
\item \textbf{Case B}: Squared exponential covariance with scaling parameter
$l=0.3$ with added random noise with $\sigma_{w}=0.01$. Covariance:
$R\left(\tau\right)=\exp\left(-\tau^{2}/l^{2}\right)+\sigma_{w}^{2}\delta\left(\tau\right)$.
Addresses \#2, \#4.
\item \textbf{Case C}: SDE(4) process with roots $\text{\ensuremath{\left\{  -0.10\pm2\pi\cdot0.25i,-0.5\pm2\pi\cdot1.5im\right\} } }$.
Addresses \#1, \#3.
\item \textbf{White noise}: A temporally uncorrelated process. Covariance
function $R\left(\tau\right)=\delta\left(\tau\right)$, where $\delta\left(0\right)=1$,
and $0$ elsewhere. Addresses: \#1.
\end{itemize}
Non-uniform sampling times are generated by drawing $N=200$ time
intervals from a Poisson distribution with average sampling interval
$T_{av}=1$. Samples $\mathbf{y}$ from a process $P$ are drawn for
the resulting sampling times $\mathbf{t}$. In this way, $S=50$ time
series $\left\{ \mathbf{t},\mathbf{y}\right\} $ are generated for
each process. For the simulated data, SDE(8) models are estimated.
The results of the simulation experiments are discussed in the subsequent
sections.

\subsection{Kullback-Leibler Discrepancy}

Our objective is to accurately characterize the second-order moments
of a random process, which we quantify using the The Kullback-Leibler
Discrepancy (KLD). The KLD has a number of desirable properties. First,
it has units that are statistically meaningful. For an unbiased estimate
of a $d$-dimensional parameter $\theta$ that achieves the Cramér-Rao
lower bound, the expected value of the KLD is asymptotically equal
to $d/2$: $\textrm{E}\left[D\left(\hat{\mathbf{\theta}}\mathbf{\Vert\mathbf{\theta}}\right)\right]=d/2$.
For the SDE($p$) model we estimate $p+1$ parameters (adding 1 to
$p$ for estimation of $\sigma_{\epsilon}$), yielding:
\begin{equation}
\textrm{E}\left[D\left(\hat{\mathbf{\theta}}\mathbf{\Vert\mathbf{\theta}}\right)\right]=\left(p+1\right)/2.\label{eq:expect-kld}
\end{equation}
 A second desirable property is that, for time series models, the
KLD is asymptotically equivalent to the spectral distortion (Root
Mean Square Error (RMSE) of the log power spectrum) and the normalized
one-step ahead prediction error. See e.g. \cite{broersen2006automatic}
for further background on these properties.

The KLD for a zero-mean multivariate Gaussian model for a random vector
$\mathbf{y}$ with covariance matrix $\hat{\Sigma}$ with respect
to the true zero-mean distribution with covariance matrix $\Sigma$
is given by:
\begin{equation}
D\left(\Sigma\Vert\hat{\Sigma}\right)=\frac{1}{2}\left(\mathrm{tr}\left(\hat{\Sigma}^{-1}\Sigma-I\right)-\log\text{\ensuremath{\left(\left|\Sigma\right|/\hat{\left|\Sigma\right|}\right)}}\right)\label{eq:kld}
\end{equation}
For SDE models, $\mathbf{y}$ is the vector of time series observations
at times $\mathbf{t}$. The covariance matrix is computed using the
covariance function from eq. \ref{eq:autocov}. The choice of time
steps $\mathbf{t}$ determines the time scale at which we evaluate
the process.

One value for $\mathbf{t}$ is the original time points of the dataset
to which the estimated model is fitted. The resulting KLD is referred
to as $D_{o}$. A second value for $\mathbf{t}$ is a regularly spaced
grid at interval $T$, referred to as $D_{T}.$ The value of $T$
corresponds to the time scale of interest at which we evaluate the
second order moments. In the frequency domain, the corresponds to
evaluating the power spectrum for frequencies up to the corresponding
Nyquist frequency, $f=1/2T$.

The KLD does not suffer from some problems associated with some alternative
ways to evaluate estimates:
\begin{itemize}
\item \textbf{Look for the ``correct'' or ``actual'' order}: Practical
processes typically cannot be described exactly by a finite order
model. Even if such a finite order model would exist, estimating a
model of this order may not result in the most accurate estimate due
to estimation errors.
\item \textbf{RMSE of estimated SDE coefficients or autocovariance}:\textbf{
}A small change coefficients or autocovariance values can result in
a completely different process, e.g. changing from stable to unstable
(coefficients), or positive-definite (valid) to not positive-definite
(invalid) for autocovariances \cite{rehfeld2011comparison}.
\end{itemize}
The base implementation of the KLD is computationally expensive. If
required, a more efficient asymptotic expression can be derived specifically
for SDE models. We do not elaborate on this here, because this computation
is only used to evaluate model performance in simulations. It not
part of the estimation algorithm, and so it will not increase computation
times for the end user. Furthermore, the generic expression allows
evaluation of non-SDE processes such as the squared exponential covariance
function.

\section{\label{sec:truncation}Data-driven model truncation}

In this section we describe the phenomenon of parameter divergence
in SDE parameter estimation, and how it can be exploited to reduce
the cost of model overfit.

\subsection{Model truncation for the SDE(1) model}

As parameters are optimized to maximize the likelihood, parameter
estimates can diverge to infinity, resulting in numerical problems
in the likelihood computation \cite{jones1981fitcar}. Also, it has
been reported that not all AR models have a continuous-time counterpart
\cite{soderstrom1991contarma}. In this section, we show how these
phenomena are closely related through a theoretical analysis of the
the SDE(1) model. Furthermore, we show how this phenomenon ultimately
results in more accurate estimates.

The SDE(1) model is an important model for many applications. It is
also known in the literature as the Ornstein-Uhlenbeck process and
is a special case of the Matérn kernel \cite{rasmussen2003gaussian}.
For data regularly sampled at interval $T$, the maximum likelihood
estimate of the SDE(1) parameter $\hat{a}$ can be computed analytically
\cite{sarkka2019sde}:

\[
\hat{a}=\frac{1}{T}\log\left[\hat{\alpha}\right]
\]
where $\hat{\alpha}$ is the estimated AR(1) parameter. For a white
noise process, the estimated AR(1) parameter is distributed symmetrically
around $\hat{\alpha}=0$ with standard deviation $1/\sqrt{N}$, where
$N$ is the number of observations \cite{priestley1981spectral}.
Hence, $\hat{\alpha}$ is negative for 50\% of signals. In this case,
the AR(1) process has no continuous-time counterpart.

It can be shown that, under these conditions, the likelihood monotonically
increases for $a\rightarrow\infty$. In this limit, the SDE(1) model
is equivalent to the white noise model. In practice, the true model
order is unknown, and so it is critical that an estimator returns
an accurate estimate under these circumstances. This is achieved using
data-driven model truncation. With model trunction, the estimation
algorithm for an SDE($p$) model can return a lower SDE($p'$) model
if the likelihood indicates that the lower order model fits better
to the data. For the SDE(1) case, this amounts in returning an SDE(0)
or white noise model when $\hat{\alpha}<0$. We discuss model truncation
for higher order models in the next section.

\subsection{Model truncation for higher order models}

As described in section \ref{subsec:Initialization}, SDE($p$) models
are estimated incrementally. If parameter divergence occurs, the model
returned by the optimization procedure has large but finite values
because of the limits introduced on parameter values. In this case,
the lower order model will have a larger likelihood, and is consequently
the final maximum likelihood estimate. While this phenomena can be
easily analyzed for the SDE(1) model estimated from white noise, this
is an important phenomena more generally, as it occurs whenever the
order of the estimated model is greater than the true model order.
Since the true model order is unknown for experimental data, it is
desirable to estimate a high order model, so that a wide range of
processes can be represented. Model truncation reduces the cost of
overfit, i.e., the statistical estimation error induced by estimation
SDE($p$) models where $p$ exceeds the true model order.

\subsection{KLD reduction achieved in white noise}

\begin{figure}[tbh]
\includegraphics[width=1\columnwidth]{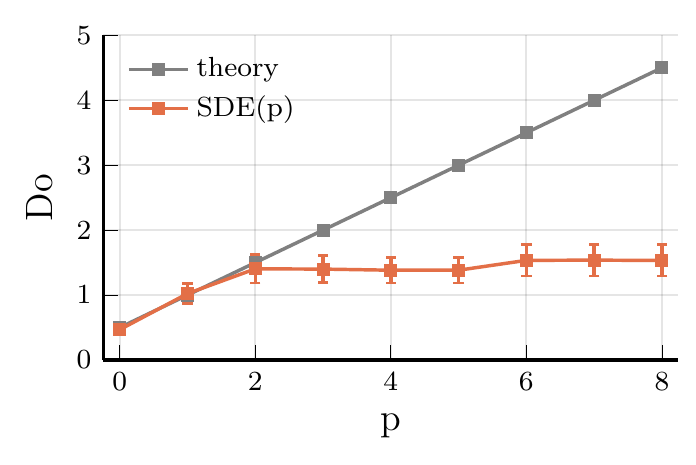}\caption{\label{fig:whitenoise}Kullback-Leibler Discrepancy (KLD) $D_{o}$
as a function of SDE model order $p$ estimated from non-uniformly
sampled white noise. Because of data-driven model truncation, the
KLD observed in simulation experiments for estimated SDE($p$) models
is below the theoretical prediction for the KLD.}
\end{figure}

Adding to the theoretical analysis of the SDE(1) case, we quantify
the improvement that is achieved with data-driven model truncation
in a simulation experiment where SDE models are estimated from a non-uniformly
sampled white noise process. The Kullback-Leibler discrepancy $D_{o}$
as a function of model order is given in figure \ref{fig:whitenoise},
along with the theoretical expected value for the KLD $D_{o}$ from
eq \ref{eq:expect-kld}. We use $D_{o}$ here, because it uses the
same time vector $\mathbf{t}$ as the likelihood. Therefore, we can
use the theoretical expectation (\ref{eq:expect-kld}).

As expected, we observe a significant reduction in the error for higher
order models due to model truncation. Because the theoretical expression
is accurate for the discrete-time AR models, these results also quantify
the error reduction compared to AR models.

\section{\label{sec:performance}Estimator performance}

\begin{figure}[tbh]
\includegraphics[width=1\columnwidth]{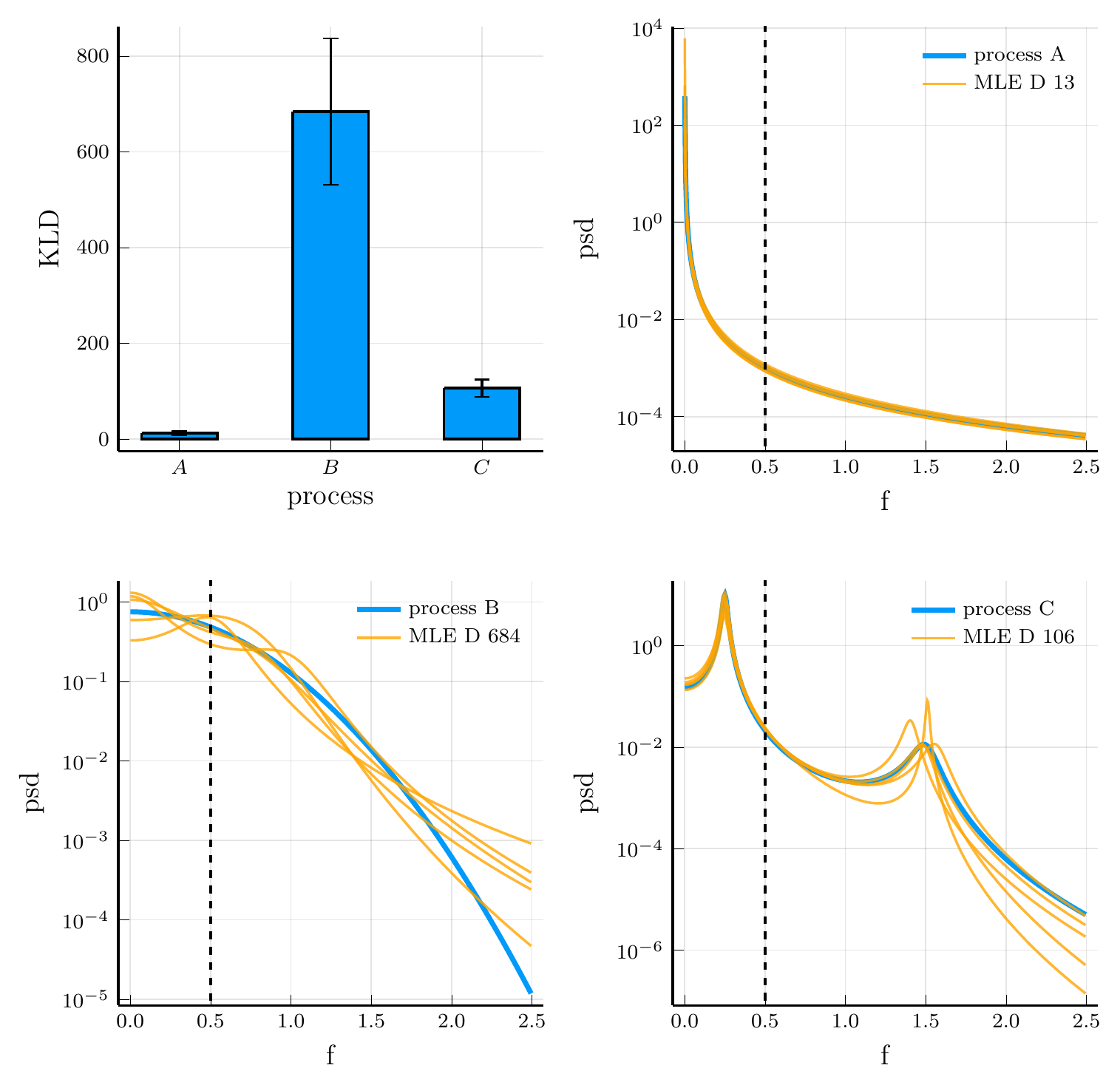}\caption{\label{fig:mle-accuracy}Accuracy of SDE(8) models estimated from
simulated data, sampled non-uniformly at an average sampling time
of $T_{av}=1$ ($S=50$ simulation runs for each case A, B and C).\textbf{
Left top:} Kullback-Leibler Discrepancy at time interval $T=0.2$
($D_{0.2}$), averaged over all simulation runs. \textbf{Remaining
graphs: }Representative sample spectral estimates computed from the
SDE(8) estimates, compared to the true power spectrum up a frequency
of $1/2T=2.5$. Accurate estimates are achieved across the entire
frequency range, going well beyond the Nyquist frequency for the average
sampling interval $f_{N,Tav}=0.5$, which is indicated by the black
dashed line.}
\end{figure}

In this section, we discuss estimator performance across the test
cases A, B and C introduced in section \ref{sec:simulations}. Figure
\ref{fig:mle-accuracy} shows the average KLD at interval $T=0.2$,
$D_{0.2}$, as well as sample spectral estimates compared to the true
spectrum up to the Nyquist frequency corresponding to the KLD interval:
$f=1/2T=2.5$. Note that we use a KLD at a time interval $T$ that
is considerably shorter than the average sampling interval $T_{av}=1$.
This allows us to quantify the estimator capability to characterize
the process at short time scales, or, in the frequency domain, up
to frequencies beyond the Nyquist rate corresponding the the average
sampling time: $f_{N,Tav}=0.5$. Estimates are shown for the SDE(8)
ML estimate obtained using the roots parameterization ($\mathbf{r}$),
referred to as the ``MLE root'' estimate.

We conclude that this estimator reliably estimates the power spectrum
for the test processes, which cover all of the challenging conditions
listed in section \ref{subsec:Test-processes}. Accurate estimates
can be obtained well above $f_{N,Tav}$. The squared exponential case
is the most challenging case, because of the joint occurrence of model
misfit and a high dynamic range. For this case, estimates are less
accurate across the entire frequency range because of statistical
estimation errors.

We specifically draw attention to the absence of large erroneous peaks
in the spectral estimates, as they have been reported previously in
the literature, including a study of the autoregressive (AR) ML estimator
for similar test cases \cite{dewaele2018kernel}. The absence of these
peaks in the SDE estimate is a consequence of the data-driven model
truncation introduced in section \ref{sec:truncation}, and results
in substantially more accurate spectral estimates. Compared to \cite{dewaele2018kernel},
the model accuracy is much improved for cases A and B. Only in case
C, the AR ML estimate in \cite{dewaele2018kernel} is more accurate.
However, for this case, a model of the true order (AR(4)) was estimated
instead of an AR(8) model, explaining the absence of erroneous peaks.
In general, we cannot assume knowledge of the true model order of
the generating process for a given experimental dataset.

Given the remarkable success in suppressing erroneous peaks compared
to previous work, the relevance of case C is to show the capability
of the estimator to accurately estimate true spectral peaks at a high
frequency. Here, our work has a key advantage over the algorithm proposed
in \cite{broersen2009spurious}, in which \emph{all} high frequency
AR roots are eliminated.

\begin{figure}[htbp]
\includegraphics[width=1\columnwidth]{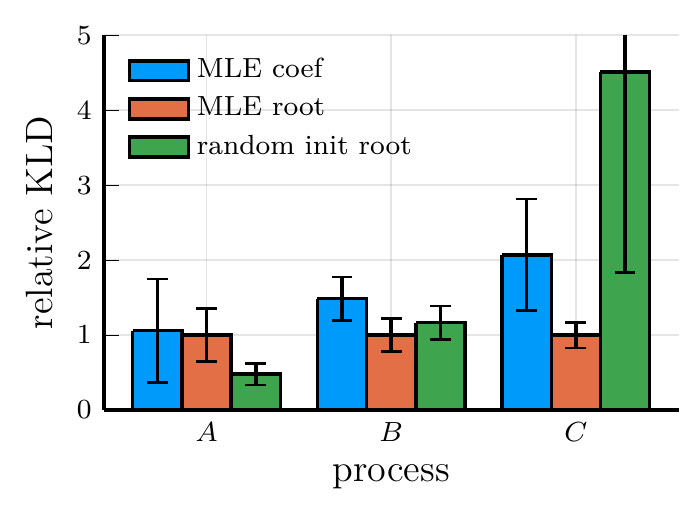}\caption{\label{fig:cmp-estimators}Comparison of the estimation accuracy of
3 alternative SDE parameter estimators. The reported values are the
Kullback-Leibler Discrepancy $D_{0.2}$ relative to the ``MLE root''
estimate.}
\end{figure}

Figure \ref{fig:cmp-estimators} compares the accuracy of the ``MLE
root'' estimator to two alternative estimators: the coefficient parameterization
``MLE coef'' and random root initiation (``random init root'').
While all estimators perform well for the basic SDE(1) case A, the
``MLE coef'' and ``random init root'' estimators have reduced
quality for the more complex cases B and C, that require accurate
higher order SDE estimates for accurate results.

For the ``MLE coef'' estimator, the reduced quality can be explained
from the fact that the likelihood computation is less numerically
stable. Also, for models of order $p>3$, the search space of positive
coefficients also contains non-stationary models. The performance
degradation is greatest for the ``random init root'' estimator,
with an increase of a factor of 4 case C. While random initiation
converges to a good solution for many simulations, it occasionally
fails to find a good optimum resulting in a very large error. This
performance degradation is expected to increase further with increasing
model order. 

\section{\label{sec:rainfall}Analysis of monsoon rainfall variability}

Long-term variability in monsoon rainfall is studied using radiometric-dated,
speleothem oxygen isotope $\delta^{18}O$ records \cite{sinha2015trends}.
The data is intrinsically irregularly sampled, because it is formed
by natural deposition rather than experimenter controlled sampling.
For the same reason, irregular sampling occurs for many other long-term
climate records as well, e.g. ice core data \cite{petit1999climate}.

The average sampling rate of speleothem data depends on the measurement
location. The current dataset is suitable for algorithm benchmarking
because it has a higher average sampling rate than datasets collected
from other locations. This allows us to study algorithm performance
for different sampling rates, by subsampling the original data, and
comparing the results to estimates obtained from the full dataset.
Also, the dataset is publicly available as supplementary material
to \cite{sinha2015trends} for reproducibility of results. The oxygen
isotope data consists of $N=1848$ irregularly sampled observations
of $\delta^{18}O$ anomalies over a time span of $2147$ years, resulting
in average sampling interval of $T_{0}=1.16$ years.

\begin{figure}[tbh]
\includegraphics[width=1\columnwidth]{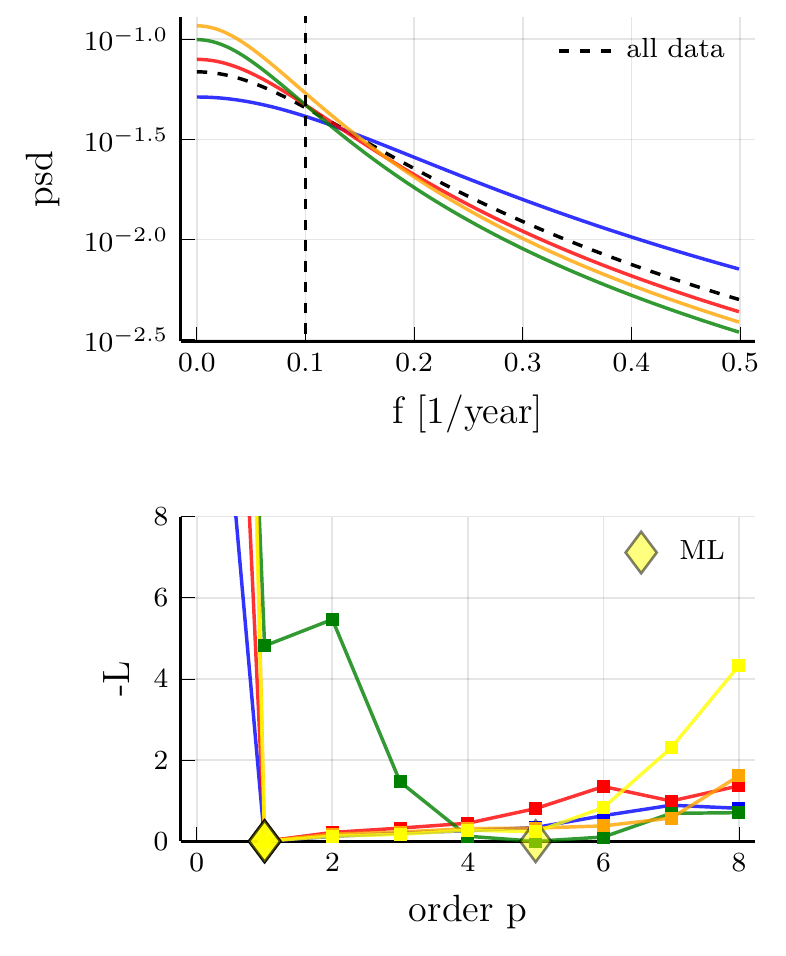}\caption{\label{fig:monsoon-psd}\textbf{Top:} Power spectra estimated from
detrended speleothem oxygen isotope $\delta^{18}O$ data using SDE(8)
estimates. Estimates are based either on all data (black dashed line),
or on subsets of data (colored lines) that are randomly subsampled
at an average sampling interval $T=5$ years. While based on a much
smaller dataset, the estimates from subsampled data remain accurate.
\textbf{Bottom}: Model fit (negative log likelihood $L$) as a function
of model order for the models estimated from subsampled data, with
colors corresponding to the same subsets as in the top spectrum plots.
The yellow diamond indicates the order of the ML estimate after data-driven
model truncation.}
\end{figure}

We estimate SDE(8) models from detrended $\delta^{18}O$ data. A reference
estimate is obtained using the complete dataset. To emulate a lower
sampling rate, we then estimate models from $5$ random subsets of
the original data, at an average sampling interval of $T_{av}=5$
years. The resulting estimated spectra and model fit are given in
figure \ref{fig:monsoon-psd}. The reference SDE(8) model is truncated
to an SDE(1) model. Models estimated from the subsampled data similarly
exploit model truncation to achieve accurate spectral estimates, avoiding
erroneous peaks in the estimate, despite a much lower sampling frequency
compared to the full dataset. The maximum likelihood is achieved at
either order $p=1$ (4 out of 5 subsets) or $p=5$ (1/5) (see figure
\ref{fig:monsoon-psd}, bottom).

\section{\label{sec:conclusions}Concluding remarks}

The proposed SDE-based method for spectral estimation from non-uniformly
sampled data provides a more accurate estimate than existing methods.
This is achieved using more accurate initialization combined with
data-driven model truncation. We have shown the performance of this
estimator in simulation experiments, and by application of the algorithm
to experimental rainfall variability data.

Further advances can be achieved through model regularization. One
way to achieve regularization is by means of order selection. Novel
order selection criteria can be developed based on the reported behavior
of SDE estimators in figure \ref{fig:whitenoise}. This behavior deviates
significantly from the theoretical behavior on which criteria such
as the Akaike Information Criterion are based.

Alternatively, regularization can be achieved by introducing informative
priors in a Bayesian approach. This has shown promise in discrete
time estimation, albeit at the cost of a considerably higher computational
load \cite{dewaele2018kernel}. Given the results in ML estimation,
we expect that a continuous-time model will also produce superior
results for Bayesian estimation.

An additional benefit of a Bayesian approach is that it has the flexibility
to produce a posterior summary that is relevant to a particular quantity
of interest. This may be exploited to produce an accurate spectral
estimate for a frequency range of interest, independent of the (average)
sampling frequency. To accommodate Bayesian inference, the code accompanying
this paper can compute posterior samples for the SDE model parameters
using Hamiltonian Monte Carlo sampling.


\end{document}